\documentclass[11pt]{article}
\usepackage{coling2020}
\usepackage{times}
\usepackage{url}
\usepackage{latexsym}

\usepackage[T1]{fontenc}
\usepackage{graphicx}
\usepackage{subfigure}
\usepackage{amsmath,amssymb}

\usepackage{algorithm} 
\usepackage{algorithmic} 
\usepackage{multirow} 
\usepackage{amsmath} 
\usepackage{xcolor}

\usepackage{color}
\usepackage{tikz}
\usepackage{pgfplots}
\usepackage{booktabs}

\colingfinalcopy 

\title{Dynamic Curriculum Learning for Low-Resource Neural Machine Translation}

\author{
  Chen Xu\textsuperscript{1}, 
  Bojie Hu\textsuperscript{2}, 
  Yufan Jiang\textsuperscript{3}, 
  Kai Feng\textsuperscript{1}, 
  Zeyang Wang\textsuperscript{1}, \\
  \textbf{Shen Huang\textsuperscript{2}, 
  {Qi Ju\textsuperscript{2}}\thanks{\ \ Corresponding author} \ , 
  Tong Xiao\textsuperscript{1,4}, 
  Jingbo Zhu\textsuperscript{1,4$\ast$ }} \\
  \textsuperscript{1}NLP Lab, School of Computer Science and Engineering, \\
   Northeastern University, Shenyang, China \\
  \textsuperscript{2}Tencent Minority-Mandarin Translation \\
  \textsuperscript{3}Smart Platform Product Department of Tencent Inc., China \\
  \textsuperscript{4}NiuTrans Research, Shenyang, China \\
  {\tt xuchenneu@163.com, \{xiaotong,zhujingbo\}@mail.neu.edu.cn} \\
  {\tt \{bojiehu, garyyfjiang, springhuang, damonju\}@tencent.com} \\
}
\date{}

\begin{document}
\maketitle
\begin{abstract}
  
  Large amounts of data has made neural machine translation (NMT) a big success in recent years. 
  But it is still a challenge if we train these models on small-scale corpora.
  In this case, the way of using data appears to be more important. 
  Here, we investigate the effective use of training data for low-resource NMT.
  In particular, we propose a dynamic curriculum learning (DCL) method to reorder training samples in training.
  Unlike previous work, we do not use a static scoring function for reordering.
  Instead, the order of training samples is dynamically determined in two ways - loss decline and model competence.
  This eases training by highlighting easy samples that the current model has enough competence to learn. 
  We test our DCL method in a Transformer-based system.
  Experimental results show that DCL outperforms several strong baselines on three low-resource machine translation benchmarks and different sized data of WMT'16 En-De.

\end{abstract}

\section{Introduction}

\blfootnote{
    %
    %
    %
    %
    \hspace{-0.65cm}  
    This work is licensed under a Creative Commons 
    Attribution 4.0 International Licence.
    Licence details:
    \url{http://creativecommons.org/licenses/by/4.0/}.
    
    
}

  Recently, neural machine translation (NMT) has demonstrated impressive performance improvements and became the de-facto standard \cite{Sutskever:nips2014,Bahdanau:ICLR2015,yonghuiwu:corr2017,vaswani:nips2017}. 
  However, like other neural methods, NMT is data-hungry.
  This makes it challenging when we train such a model in low-resource scenarios \cite{koehn:acl2017}.
  Researchers have developed promising approaches to low-resource NMT.
  Among these are data augmentation \cite{Sennrich:acl2016,Cheng:acl2016,Fadaee:acl2017}, transfer learning \cite{Zoph:emnlp2016,Kocmi:wmt2018}, and pre-trained models \cite{Peters:naacl2018,Devlin:naacl2019}.
  But these approaches rely on external data other than bi-text.
  To date, it is rare to see work on the effective use of bilingual data for low-resource NMT.

  In general, the way of feeding samples plays an important role in training neural models.
  A good instance is that it is popular to shuffle the input data for robust training in state-of-the-art systems.
  More systematic studies on this issue can be found in recent papers \cite{Bengio:icml2009,Kumar:nips2010,Shrivastava:cvpr2016}.
  For example, \newcite{Arpit:icml2017} have pointed out that deep neural networks tend to prioritize learning ``easy'' samples first.
  This agrees with the idea of curriculum learning \cite{Bengio:icml2009} in that an easy-to-hard learning strategy can yield better convergence for training.

  In NMT, curriculum learning is not new.
  Several research groups have applied it to large-scale translation tasks although few of them discuss the issue in a low-resource setup \cite{Zhang:corr2018,Platanios:naacl2019,Liu:acl2020}.
  The first question here is how to define the ``difficulty'' of a training sample.
  Previous work resorts to functions that produce a difficulty score for each training sample.
  This score is then used to reorder samples before training.
  But the methods of this type enforce a static scoring strategy and somehow disagrees with the fact that the sample difficulty might be changing when the model is updated during training.
  Another assumption behind curriculum learning is that the difficulty of a sample should fit the competence of the model we are training.
  Researchers have implicitly modeled this issue by hand-crafted curriculum schedules \cite{Zhang:corr2018} or simple functions \cite{Platanios:naacl2019}, whereas there has no in-depth discussion on it yet.

  In this paper, we continue the line of research on curriculum learning in low-resource NMT.
  We propose a dynamic curriculum learning (DCL) method to address the problems discussed above.
  The novelty of DCL is two-fold.
  First, we define the difficulty of a sample to be the decline of loss (i.e., negative log-likelihood).
  In this way, we can measure how hard a sentence can be translated via the real objective used in training.
  Apart from this, the DCL method explicitly estimates the model competence once the model is updated, so that one can select samples that the newly-updated model has enough competence to learn.
  
  DCL is general and applicable to any NMT system.
  In this work, we test it in a Transformer-based system on three low-resource MT benchmarks and different sized data selected from the WMT'16 En-De task.
  Experimental results show that our system outperforms the strong baselines and several curriculum learning-based counterparts.
  
\section{Related work}

\subsection{Low-Resource NMT}

  \newcite{koehn:acl2017} show that NMT systems result in worse translation performance in low-resource scenarios.
  Researchers have developed promising approaches to this problem which mainly focus on introducing external knowledge to improve low-resource NMT performance.
  Data augmentation \cite{Sennrich:acl2016,Cheng:acl2016,Fadaee:acl2017} alleviates this problem by generating pseudo parallel data. 
  A large amount of auxiliary parallel corpus from other related language pairs can be used to pre-train model parameters and transfer to target language pair \cite{Zoph:emnlp2016,Chen:acl2017,Gu:naacl2018,Gu:emnlp2018,Kocmi:wmt2018}.
  Pre-trained language models trained with a large amount of monolingual data \cite{Peters:naacl2018,Devlin:naacl2019} improve the quality of NMT model significantly \cite{Clinchant:wngt2019,Yang:aaai2020,Zhu:iclr2020}.

  However, these approaches rely on a large number of external resources or conditions, e.g., the auxiliary parallel corpus related to the source or target language, or a large amount of monolingual data. 
  \newcite{Sennrich:acl2019} demonstrate the competitive NMT model can be trained with the appropriate hyperparameters in low-resource scenarios without any external resources.
  This is consistent with our motivation.
  The difference lies in that they focus on the model settings, and we explore the training strategy which utilizes bilingual data effectively for low-resource NMT.
  
\subsection{Curriculum Learning}

  Curriculum learning \cite{Bengio:icml2009} is motivated by the learning strategy of biological organisms which orders the training samples in an easy-to-hard manner \cite{Elman:cognition1993}.
  Benefited from organized training, the neural network explores harder samples effectively utilizing the previous knowledge learned from easier samples.
  \newcite{Weinshall:icml2018} demonstrate curriculum learning speeds up the learning process, especially at the beginning of training.
  Curriculum learning has been applied to several tasks, including language modeling \cite{Bengio:icml2009}, image classification \cite{Weinshall:icml2018}, and human attribute analysis \cite{Wang:iccv2019}.

  Curriculum learning has recently shown to train large-scale translation tasks efficiently and effectively by controlling the way of feeding samples.
  \newcite{Kocmi:ranlp2017} construct mini-batch contains sentences similar in length and linguistic phenomena, then organize the order by increased complexity in one epoch.
  \newcite{Zhang:corr2018} group the training samples into shards based on model-based and linguistic difficulty criteria, then train with hand-crafted curriculum schedules.
  \newcite{Platanios:naacl2019} propose competence-based curriculum learning that select training samples based on sample difficulty and model competence.
  \newcite{Kumar:naacl2019} use reinforcement learning to learn the curriculum automatically.
  \newcite{Liu:acl2020} propose a norm-based curriculum learning method based on the norm of word embedding to improve the efficiency of training an NMT system.
  \newcite{Zhou:acl2020} propose uncertainty-aware curriculum learning.
  To the best of our knowledge, this is the first comprehensive discussion of curriculum learning in a low-resource setup.

  On the other hand, curriculum learning is similar to data selection and data sampling methods.
  More similar work is that \newcite{Wang:acl2018} propose a dynamic sampling method that calculates the decline of loss during training to improve the NMT training efficiency.
  They start training from the full training set and then gradually decrease. 
  This is contrary to the idea of curriculum learning.


\section{Problem Definition}

  Let $D_{train}$ be the training corpus and $|D_{train}|$ be the corpus size.
  $\mathbf{s} = (\mathbf{x}, \mathbf{y})$ is a training sample in $D_{train}$, where $\mathbf{x}$ is the source sentence and $\mathbf{y}$ is the target sentence.
  NMT systems learn a conditional probability $P(\mathbf{y}|\mathbf{x})$:
  \begin{eqnarray}
    \begin{aligned}
      P(\mathbf{y}|\mathbf{x}; \boldsymbol{\theta}) &= \prod_{i=1}^{|\mathbf{y}|} P(\mathbf{y}_i|\mathbf{x}, \mathbf{y}_{<i}; \boldsymbol{\theta})
    \end{aligned}	 
  \end{eqnarray}
  where $|\mathbf{y}|$ is the length of $\mathbf{y}$, $\boldsymbol{\theta}$ is a set of model parameters.
  The training objective is to seek the optimal parameters $\boldsymbol{\hat{\theta}}$ by minimizing the negative log-likelihood (NLL) of the training set:
  \begin{eqnarray}
    \begin{aligned}
      \boldsymbol{\hat{\theta}} &= \mathop{\arg\min}_{\boldsymbol{\theta}} \sum_{\mathbf{s} \in D_{train}} -{\rm log} P(\mathbf{y}|\mathbf{x}; \boldsymbol{\theta}) \\
               &= \mathop{\arg\min}_{\boldsymbol{\theta}} \sum_{\mathbf{s} \in D_{train}} \sum_{i=1}^{|\mathbf{y}|} -{\rm log} P(\mathbf{y}_i|\mathbf{x}, \mathbf{y}_{<i}; \boldsymbol{\theta})
  \end{aligned}
  \label{eq2}
  \end{eqnarray}

  Our objective is to learn better model parameters by curriculum learning in low-resource NMT.
  We decompose the whole training process into multiple phases $T = (t^0, t^1, t^2, \dots, t^k)$\footnote{In the standard NMT, every epoch is a phase.}.
  For every phase $t$, the sub-optimal process can be viewed as:
  \begin{eqnarray}
    \begin{aligned}
      \boldsymbol{\theta^{t+1}} &= \mathop{\arg\min}_{\boldsymbol{\theta^t}} \sum_{\mathbf{s} \in D_{train}^t} -{\rm log} P(\mathbf{y}|\mathbf{x}; \boldsymbol{\theta^{t}})
  \end{aligned}
  \label{eq3}
  \end{eqnarray}
  where $\boldsymbol{\theta^{t}}$ is the model parameter at phase $t$.
  Two sub-questions \cite{Platanios:naacl2019} in curriculum learning are separated to determine the training data $D_{train}^t$:
  
  \begin{itemize}
    \item \textbf{Sample Difficulty}. How to measure the difficulty of a training sample with a quantified value?
    \item \textbf{Model Competence}. How to estimate the competence of the model to arrange training data that model can learn effectively?
  \end{itemize}

  Previous work enforces a static scoring strategy to measure sample difficulty and encourage simple functions to estimate the model competence.
  In this way, the training data at each phase is pre-determined before training.
  But in fact, sample difficulty and model competence are not independent with the current model parameters.

  A natural idea is to re-arrange the curriculum once the model is updated, so that we can select training data with appropriate difficulty for current training.
  We discuss these two questions in-depth in the following section.

\section{Dynamic Curriculum Learning}

  \label{DCL}
  We propose a dynamic curriculum learning method to reorder training samples in training.
  We determine the order of training samples dynamically, rather than using a static scoring for reordering.
  Besides, we propose a batching method to reduce gradient noise.

  \subsection{Sample Difficulty}
  \label{Section4.1}

  Equation \ref{eq2} shows that the training objective of NMT is to minimize the loss (i.e., negative log-likelihood) of the training set.
  For a training sample $\mathbf{s}=(\mathbf{x}, \mathbf{y})$ at phase $t$, the loss is calculated as:
  \begin{eqnarray}
    \begin{aligned}
      l(\mathbf{s}; \boldsymbol{\theta}^t) = -{\rm log} P(\mathbf{y}|\mathbf{x}; \boldsymbol{\theta}^t)
    \end{aligned}
    \label{predicted_loss}
  \end{eqnarray}

  The NMT model generally translates sentences with lower loss better.
  Based on this idea, \newcite{Zhang:corr2018} define the difficulty as the probability of the top-1 translation candidates generated by a pre-trained NMT model.
  While translation probability related to the training objective represents the sample difficulty accurately compared with the heuristic metrics, it still suffers from the problem of static scoring.
  Therefore, a natural idea is calculating the loss of training samples dynamically to measure its difficulty.
  To this end, we evaluate the loss of all the training samples with the fixed model parameters before each training phase:
  \begin{eqnarray}
    \begin{aligned}
      d(\mathbf{s}; \boldsymbol{\theta}^t) = l(\mathbf{s}; \boldsymbol{\theta}^t)
    \end{aligned}
    \label{dynamic_loss}
  \end{eqnarray}
  where $d(\mathbf{s}; \boldsymbol{\theta}^t)$ is the difficulty of the sample $\mathbf{s}$ at phase $t$.

  While the loss shows the level that the current model can handle this sample, there also suffers from two drawbacks \cite{Wang:acl2018}.
  First, the loss of a sample may be large at the initial phase but easy to decrease rapidly after a few phases.
  Second, one sentence with small loss may have no space to further decrease and training model with these sentences iteratively may lead to overfitting.

  Therefore, we define the difficulty of a sample to be the decline of loss.
  In this way, we take into account the model change between the previous phase and the current phase.
  The decline-based sample difficulty is measured as:
  \begin{eqnarray}
    \begin{aligned}
      d(\mathbf{s}; \boldsymbol{\theta}^t, \boldsymbol{\theta}^{t-a}) = \begin{cases}
        \ \ \ \ \ \ \ l(\mathbf{s}; \boldsymbol{\theta}^t) & t < a \\
        \\
        \displaystyle -\frac{l(\mathbf{s}; \boldsymbol{\theta}^{t-a}) - l(\mathbf{s}; \boldsymbol{\theta}^t)}{l(\mathbf{s}; \boldsymbol{\theta}^{t-a})} & t \ge a
      \end{cases}
    \end{aligned}
    \label{ratio}
  \end{eqnarray}
  where $a \ge 1$ represents we compare the loss decline at phase $t-a$ and phase $t$.

  Based on this difficulty metric, the sentence with low difficulty indicates the NMT model improves the predicted accuracy of its translation result significantly.
  Therefore, it is more likely to learn better in the next phase.
  On the contrary, the sentence with high difficulty indicates the current NMT model does not have enough competence to handle it and may wait to be learned at a later phase.

  \subsection{Model Competence}

  \newcite{Platanios:naacl2019} propose a competence-based curriculum learning framework which defines the model competence $c(t) \in (0, 1]$ at training step $t$ by simple functional forms:

  \begin{eqnarray}
    \begin{aligned}
      c(t) = {\rm min} (1, \sqrt[p]{\frac{t}{T}(1-c_0^p) + c_0^p})
    \end{aligned}	 
    \label{func}
  \end{eqnarray}
  where $c_0 \ge 0$ is the initial competence, $p$ is the coefficient to control the curriculum schedule.
  Competence is seen as a linear function when $p=1$ and the harder samples increase by a constant amount, square root function when $p=2$ and the harder samples increase faster in the early phases and slower in the later phases.

  However, these intuitive functions might not be universal for model competence.
  There is a gap between the high-resource and low-resource tasks.
  Limited by a small amount of data, the performance of the NMT model improves slowly at the early phases (see Section \ref{sec_curve}).
  
  In this paper, we propose a dynamic estimation method, which measures model competence at every phase based on the performance of the development set.
  While the loss of development set is an optional method, the sentence-level evaluation metric BLEU \cite{Papineni:acl2002} presents more superiorities \cite{Shen:acl2016}.
  In this way, the model competence avoids the prior hypothesis of the training process and is related to the real performance on the unseen dataset.

  Specifically, we pre-train a vanilla NMT model and record the best BLEU value on the development set as curriculum length $\mathrm{BLEU}_{T}$.
  The model competence is estimated as:
  \begin{eqnarray}
    \begin{aligned}
      c(t) = {\rm min}(1, {\frac{\mathrm{BLEU}_t}{\mathrm{BLEU}_{T} * \beta} (1-c_0) + c_0})
    \end{aligned}
    \label{competence}
  \end{eqnarray}
  where $\mathrm{BLEU}_t$ is the BLEU at phase $t$, $\beta \in (0, 1]$ is a coefficient to control the curriculum speed.
  With a smaller $\beta$, the progress of curriculum learning is faster and the model can be trained on the entire training set earlier.

  We suppose that the model has weak competence to only learn well from the easiest training samples at the initial phase and gradually has enough competence to handle the entire training set $D_{train}$.
  We measure the sample difficulty and model competence before every phase, then the $|D_{train}| * c(t)$ easiest training samples are selected to train the NMT model.
  Benefited from dynamic measurement, the newly-updated model has enough competence to learn samples with the appropriate difficulties.

  \subsection{Batching}

  \newcite{Goyal:corr2017} address the optimization difficulty when training a neural network with large batches and exhibit good generalization.
  Large batches have demonstrated better performance in high-resource NMT tasks due to the lower gradient noise scale \cite{Ott:wmt2018,Popel:ling2018}.
  However, this method degrades the performance in low-resource tasks due to poorer generalization \cite{Keskar:iclr2017}.
  
  The dominant NMT batches the samples with similar lengths to speed training up \cite{Khomenko:corr2017}.
  To reduce gradient noise, we propose a batching method which batches the samples based on similar difficulty in our curriculum learning method.
  Samples with similar difficulty indicate their losses fall at a similar rate.
  It might have a stabilized gradient direction and leads to better performance.

  \begin{algorithm}[t]
    \caption{NMT with Dynamic curriculum learning}
    \begin{algorithmic}[1]

    \REQUIRE The training set $D_{train}$, the development set $D_{dev}$, \\
    \ \ \ \ \ \ the best BLEU value of a baseline model $\mathrm{BLEU}_{T}$.
    \ENSURE A NMT model with dynamic curriculum learning.
    
    \STATE $t=0$; Randomly initialize the model parameters $\boldsymbol{\theta}^0$;

    \WHILE{$t < T$}
      \STATE Evaluate the $D_{train}$ and get loss $l(\mathbf{s}; \boldsymbol{\theta}^t)$ of every training sample $\mathbf{s}$;
      \FORALL{$\mathbf{s}$ in $D_{train}$}
        \STATE Measure the difficulty of $\mathbf{s}$ by Equation \ref{ratio};
      \ENDFOR 
      \STATE Sort $D_{train}$ based on the difficulty of every training sample;
      \STATE Evaluate the $D_{dev}$ and get $\mathrm{BLEU}_t$;
      \STATE Estimate the model competence $c(t)$ by Equation \ref{competence};
      \STATE Train the NMT model with the $|D_{train}| * c(t)$ easiest training samples;
      \STATE $t \leftarrow t + 1$;
    \ENDWHILE 
    \end{algorithmic}
    \label{algorithm}
  \end{algorithm}

  \subsection{Training Strategy}

  \newcite{Zhang:corr2018} define two general types of curriculum learning strategy.
  The deterministic curriculum \cite{Kocmi:ranlp2017} arranges the training samples with fixed order and performs worse due to lacking randomization.
  The probabilistic curriculum \cite{Platanios:naacl2019} generates a batch uniformly sampled from the training set based on sample difficulty and model competence.
  The latter generally works well in the previous curriculum learning methods.
  However, in our preliminary experiments, we find that the vanilla model trained by sampling performs worse or converges slower slightly than the training strategy which trains the model with the whole training set in an epoch.
  A possible reason is sampling might lead to unbalanced training because some samples are not fully trained due to sampling omission.

  Our method dynamically measures the sample difficulty and model competence at each phase, then selects a certain proportion of easier samples to train based on model competence.
  It ensures that training samples are not missed due to sampling and also retains randomization to avoid overfitting.
  Algorithm \ref{algorithm} shows the overall training procedure of our method.

\section{Experiment Setup}

  \subsection{Datasets}

  We consider three different low-resource machine translation benchmarks from IWSLT TED talk, running experiments in IWSLT'15 Chinese-English (\textbf{Zh-En}), IWSLT'15 Thai-English (\textbf{Th-En}), and IWSLT'17 English-Japanese (\textbf{En-Ja}). 
  We concatenate dev2010, tst2010, tst2011, tst2012, and tst2013 as the development set.
  We use tst2015 as the test set for Zh-En and En-Th, tst2017 for En-Ja.
  To simulate different amounts of training resources, we randomly subsample 50K/100K/300K sentence pairs from WMT'16 English-German dataset and denote them as \textbf{50K/100K/300K}.
  Furthermore, we also verify the effect of our method in the high-resource scenarios with all WMT'16 English-German training set (\textbf{4.5M}).
  We concatenate newstest2012 and newstest2013 as the development set and newstest2016 as the test set.
  Data statistics are shown in Table \ref{statistics}.

  We tokenize the Chinese sentences using NiuTrans \cite{Xiao:acl2012} word segmentation tookit and Japanese sentences using MeCab\footnote{https://taku910.github.io/mecab/}.
  For other language pairs, we apply the same tokenization using Moses scripts \cite{Koehn:acl2007}.
  We learn Byte-Pair Encoding \cite{Sennrich:acl2016b} subword segmentation with 10,000 merge operations for IWSLT datasets and 32,000 merge operations for WMT dataset.
  Especially, we learn BPE with a shared vocabulary for WMT dataset.

  \begin{table}[ht]
    \centering
    \begin{tabular}{c|c|c|c}
        \toprule
        Dataset & \# Train & \# Dev & \# Test \\
        \midrule
        IWSLT'15 En-Th & 85019 & 4904 & 756 \\
        IWSLT'15 En-Zh & 213377 & 6360 & 1080 \\
        IWSLT'17 En-Ja & 226834 & 6861 & 1452 \\
        WMT'16 En-De & 50K/100K/300K/4.5M & 6003 & 2999 \\
        \bottomrule
    \end{tabular}
    \caption{Number of sentences in each dataset.}
    \label{statistics}
  \end{table}

  \subsection{Model Settings}

  In all experiments, we use the \textit{fairseq} \cite{Ott:naacl2019}\footnote{https://github.com/pytorch/fairseq} implementation of the Transformer. 
  Inspired by \cite{Sennrich:acl2019}, we select different hyperparameters for each dataset to build a strong baseline in low-resource NMT\footnote{Universal hyperparameters lead to a weaker baseline on some low-resource data sets, even not converge. We report the results of the stronger baseline trained with tuned hyperparameters on the development set to demonstrate the effectiveness of our method.}. 
  The model hyperparameters are shown in Table \ref{hyperparameter}.
  We use the Adam optimizer \cite{Kingma:iclr2015} with $\beta_{1} = 0.9$, $\beta_{2} = 0.98$, and $\epsilon = 10^{-9}$.
  We increase the learning rate from $10^{-7}$ to $0.0005$ for IWSLT datasets and $0.0007$ for WMT dataset with linear warmup over the first $4000$ steps and decay the learning rate by inverse square root way.
  For all experiments, we use the dropout of 0.1 and label smoothing $\epsilon_{ls}=0.1$ for regularization.
  We share the embedding and the output layers of the decoder for IWSLT experiments, share all embedding for WMT experiments.

  \begin{table}[ht]
    \centering
    \begin{tabular}{c|c|c|c|c}
        \toprule
        Dataset & Embedding Size & Feed-Forward Size & Depth & Head \\
        \midrule
        Th-En & 512 & 1024 & 6 & 4 \\
        Zh-En & 512 & 1024 & 6 & 4 \\
        En-Ja & 512 & 1024 & 4 & 4 \\
        50K/100K/300K & 512 & 2048 & 4 & 8 \\
        4.5M & 512 & 2048 & 6 & 8 \\
        \bottomrule
    \end{tabular}
    \caption{Model hyperparameters for each task.}
    \label{hyperparameter}
  \end{table}

  \subsection{Training and Inference}

  We train the model with batch size consisted of $4096$ tokens except for 4.5M which uses $4096 * 8$ tokens to compare with a standard setting.
  We evaluate the development set at every epoch during training by decoding using greedy search and calculate the BLEU\footnote{https://github.com/moses-smt/mosesdecoder/tree/master/scripts/generic/multi-bleu.perl} \cite{Papineni:acl2002} score.
  We early stop training when there is no improvement of the BLEU for 10 consecutive checkpoints.
  During inference, we use beam search with the beam size of 5 and length penalty of 1 for IWSLT datasets, beam size of 4 and length penalty of 0.6 for WMT datasets.

  \subsection{Curriculum Learning Setup}

  We implement our method with \textit{fairseq} by simply modifying.
  Our DCL method measures the sample difficulty and model competence before every phase dynamically.
  While it results in extra time consumption (about $30\%$), it is acceptable for low-resource tasks.

  In all experiments, we set $a=1$ uniformly in Equation \ref{ratio} to measure the difficulty, which means the sample difficulty takes into account two adjacent phases.
  For the model competence described in Equation \ref{competence}, we record the best BLEU of the baseline model on the development set as the $\mathrm{BLEU}_{T}$.
  Although hyperparameter with careful selection can bring improvement, we set $c_0 = 0.2$ and $\beta = 0.9$ universally for all experiments.
  It means we start training with the $20\%$ easiest sentences and train the model with the whole training set when the performance achieves $90\%$ of $\mathrm{BLEU}_{T}$.
  
  We use the following notations to represent different curriculum learning strategies.
  For the sample difficulty, we compare our method with previous heuristic metrics and two other dynamic metrics: 
  
  \begin{itemize}
    \setlength{\itemsep}{0pt}
    \setlength{\parsep}{0pt}
    \setlength{\parskip}{0pt}
    \item Heuristic metrics: Source sentence length (\textbf{Length}) and source word rarity (\textbf{Rarity}) \cite{Platanios:naacl2019}.
    \item Dynamic metrics: Random difficulty value (\textbf{Random}) and the loss at the current phase (\textbf{Loss}).
    \item Our method: Loss decline between the previous phase and the current phase (\textbf{Decline}).
  \end{itemize}
  For the model competence, we experiment with the following methods:
  \begin{itemize}
  \setlength{\itemsep}{0pt}
  \setlength{\parsep}{0pt}
  \setlength{\parskip}{0pt}
  \item Functional forms: Linear (\textbf{Linear}) and square root (\textbf{Sqrt}) model competence \cite{Platanios:naacl2019}.
  \item Our method: Dynamic model competence (\textbf{DMC}) based on the performance.
  \end{itemize}
  
\section{Results}

\begin{table}[t]
  \centering
    \begin{tabular}{l|c|c|c|c|c|c|c}
      \toprule
        Method & Th-En & Zh-En & En-Ja & 50K & 100K & 300K & 4.5M \\
        \midrule   
        Baseline & 18.71 & 20.64 & 12.33 & 9.21 & 17.56 & 23.84 & 33.49 \\
        \midrule
        Length + Sqrt & 18.89 & 20.51 & 11.90 & 8.16 & 16.67 & 23.86 & 32.96 \\
        Rarity + Sqrt & 18.92 & 20.65 & 12.78 & 8.83 & 16.58 & 24.23 & 33.17 \\
        \midrule
        Random + DMC & 18.00 & 20.93 & 13.03 & 7.80 & 17.35 & 24.10 & 33.47 \\
        Loss + DMC & 18.34 & 20.76 & 12.59 & 8.46 & 17.06 & 24.35 & 33.30 \\
        \midrule
        Decline + Linear & 18.95 & 20.75 & 12.68 & 10.87 & 18.03 & 24.02 & \textbf{33.61} \\
        Decline + Sqrt & 19.16 & 21.05 & 12.80 & 10.49 & 17.86 & \textbf{24.63} & 33.52 \\
        \midrule
        Decline + DMC & 19.13 & 21.14 & 13.24 & 10.65 & 17.89 & 24.35 & 33.52 \\
        \ \ +Batching & \textbf{19.32} & \textbf{21.20} & \textbf{13.47} & \textbf{12.90} & \textbf{18.44} & 24.45 & / \\
      \bottomrule
    \end{tabular}
    \caption{BLEU scores (\%) of different methods in each tasks. 
    +Batching represents we batch the sentences with similar difficulty.}
    \label{results}
\end{table}

  Table \ref{results} summaries the experimental results with different curriculum learning methods.
  The existing curriculum learning methods (row 2 and row 3) can not improve performance stably or even degraded, which demonstrates the heuristic metrics are not helpful in low-resource scenarios.

  With the dynamic model competence, we observe the BLEU scores of Random (row 4) fluctuates on the baseline model.
  Although the difficulty is measured dynamically during training, the meaningless value is not favorable for curriculum learning.
  We also measure the difficulty with loss dynamically (row 5), which leads to a slight improvement in the larger datasets.
  However, it degrades performance in the smaller datasets (Th-En, 50K, and 100K).
  It agrees with analysis in Section \ref{Section4.1} that it is easy to fall into overfitting due to repeated training on some samples with low loss, especially in the extremely scarce datasets.

  Then, we test our proposed difficulty metric of Decline.
  With the simple competence functions (row 6 and row 7), they outperform significantly the strong baselines and previous methods.
  It demonstrates our proposed metric is of high relevance with sample difficulty for NMT than the other four metrics.
  However, the Sqrt competence function does not perform better than the Linear function in all datasets, this is not consistent with the previous conclusion \cite{Platanios:naacl2019}.
  The possible reason is that the model competence of some low-resource datasets improves slowly in the early phases.
  DMC (row 8) avoids the prior hypothesis of the performance change and achieves better or similar performance compared with the above methods.

  Finally, we batch the samples with similar difficulties\footnote{We do not do this experiment on the En-De 4.5M dataset because it is trained with large batches.} (row 9).
  While the model is trained with more training steps due to padding, it achieves further improvement over our curriculum learning method in all tasks.
  This verifies our hypothesis that the gradient of samples with similar difficulty is more stabilized.
  This is an interesting result and we will explore it in future work.

  Overall, the experimental results show our proposed method achieves better performance compared with strong baselines and several curriculum learning-based counterparts in the low-resource NMT tasks.

\section{Analysis}

  We take En-De 50K dataset which achieves the most performance improvement to analyze our method.
  Although it is sampled from WMT dataset, we think it can demonstrate the advantages of our method obviously.

  \begin{figure}[]
  \pgfplotsset{width=4.2cm,compat=1.16}
  \begin{center}
    \ref{named}
    \\
    \begin{tikzpicture}
      \begin{axis}[
        axis lines=left,
        legend columns=2,
        legend entries={Baseline\quad\quad\quad, Decline + DMC},
        legend to name=named,
        legend style={draw=none,line width=0.8pt},
        title={En-De 50K},
        xmin=0, xmax=10000,
        ymin=0, ymax=8,
        xtick={0, 5000, 10000},
        xticklabels={0, 5K, 10K},
        ytick={0, 4, 8},
        xlabel=Steps,
        ylabel=BLEU,
        scaled ticks=false,
        line width=0.8pt,
        ]
      \addplot[mark=none, color=blue] file {data/50k_baseline.txt};
      \addplot[mark=none, color=red] file {data/50k_ratio.txt};
      \end{axis}
    \end{tikzpicture}
    \begin{tikzpicture}
      \begin{axis}[
        axis lines=left,
        title={En-De 100K},
        xmin=0, xmax=20000,
        ymin=0, ymax=14,
        xtick={0, 10000, 20000},
        xticklabels={0, 10K, 20K},
        ytick={0, 7, 14},
        xlabel=Steps,
        scaled ticks=false,
        line width=0.8pt,
        ]
      \addplot[mark=none, color=blue] file {data/100k_baseline.txt};
      \addplot[mark=none, color=red] file {data/100k_ratio.txt};
      \end{axis}
    \end{tikzpicture}
    \begin{tikzpicture}
      \begin{axis}[
        axis lines=left,
        legend columns=-1,
        title={En-De 300K},
        xmin=0, xmax=50000,
        ymin=10, ymax=18,
        xtick={0, 25000, 50000},
        xticklabels={0, 25K, 50K},
        ytick={10, 14, 18},
        xlabel=Steps,
        scaled ticks=false,
        line width=0.8pt,
        ]
      \addplot[mark=none, color=blue] file {data/300k_baseline.txt};
      \addplot[mark=none, color=red] file {data/300k_ratio.txt};
      \end{axis}
    \end{tikzpicture}
    \begin{tikzpicture}
      \begin{axis}[
        axis lines=left,
        title={En-De 4.5M},
        xmin=0, xmax=100000,
        ymin=20, ymax=24,
        xtick={0, 50000, 100000},
        xticklabels={0, 50K, 100K},
        ytick={20, 22, 24},
        xlabel=Steps,
        scaled ticks=false,
        line width=0.8pt,
        ]
      \addplot[mark=none, color=blue] file {data/4.5m_baseline.txt};
      \addplot[mark=none, color=red] file {data/4.5m_ratio.txt};
      \end{axis}
    \end{tikzpicture}
  \end{center}

    \caption{BLEU scores (\%) of the development set on WMT'16 En-De dataset of different sizes.}
    \label{learning_curve}
  \end{figure}
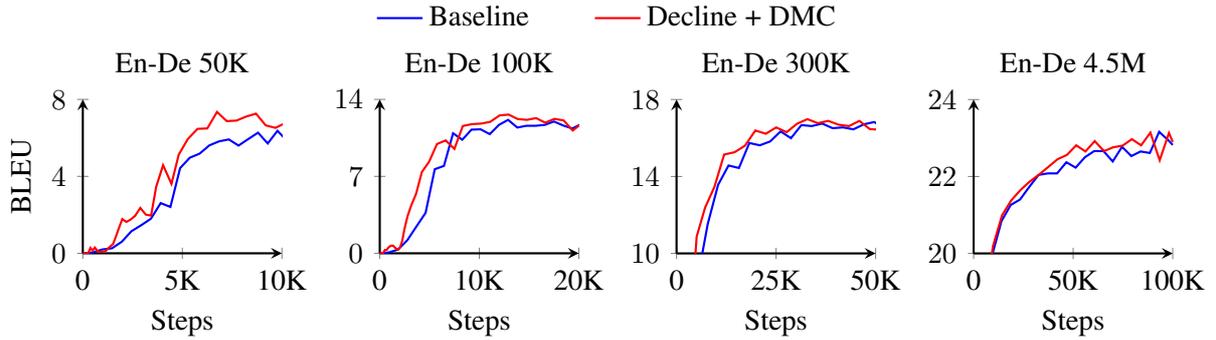

  \subsection{Learning Curve}
  \label{sec_curve}

  \begin{figure}[t]
    \begin{center}
    \ref{test}
    \end{center}
    \begin{minipage}[t]{0.3\textwidth}
      \pgfplotsset{width=5cm,compat=1.16}
        \begin{tikzpicture}
          \begin{axis}[
            axis lines=left,
            legend columns=2,
            legend entries={Baseline\quad\quad\quad, Decline + DMC},
            legend to name=test,
            legend style={
              draw=none,
              line width=0.8pt,
            },
            xmin=0, xmax=11,
            ymin=4, ymax=9,
            xtick={0, 5, 10},
            xlabel=Training Counts,
            ylabel=Loss,
            scaled ticks=false,
            line width=0.8pt,
            ]
        \addplot[mark=*, color=blue] file {data/baseline_avg_loss.txt};
        \addplot[mark=*, color=red] file {data/ratio_avg_loss.txt};
        \end{axis}
      \end{tikzpicture}
      \caption{Average loss of all the training samples during training.}
      \label{avg_loss}
    \end{minipage}
    \hspace{0.02\textwidth}
    \begin{minipage}[t]{0.3\textwidth}
      \pgfplotsset{width=5cm,compat=1.16}
        \begin{tikzpicture}
          \begin{axis}[
            axis lines=left,
            legend style={
            legend pos=south east,
            },
            legend style={line width=0.8pt},
            xmax=4.5,
            ymin=10, ymax=22,
            ytick={10, 15, 20},
            xtick={0, 2, 4},
            xticklabels={{$(0,10)$}, {$[20,30)$}, {$[40,+\infty)$}},
            xlabel=Sentence Lengths,
            ylabel=Training Counts,
            line width=0.8pt,
            scaled ticks=false,
            ]
          \addplot [sharp plot, mark=*, color=blue] coordinates {
            (0, 20) (1, 20) (2, 20) (3, 20) (4, 20)
            };
          \addplot [sharp plot, mark=*, color=red] coordinates {
            (0, 15.3) (1, 13.77) (2, 12.93) (3, 12.4) (4, 11.89)
            };
          \end{axis}
        \end{tikzpicture}
        \caption{Training counts of samples with different sentence lengths in the training set.}
        \label{training_counts}
    \end{minipage}
    \hspace{0.02\textwidth}
    \begin{minipage}[t]{0.3\textwidth}
    \pgfplotsset{width=5cm,compat=1.16}
      \begin{tikzpicture}
        \begin{axis}[
          axis lines=left,
          ymin=6, ymax=12,
          xmin=0, xmax=4.5,
          xtick={0, 2, 4},
          xticklabels={{$(0,10)$}, {$[20,30)$}, {$[40,+\infty)$}},
          xlabel=Sentence Lengths,
          ylabel=BLEU,
          line width=0.8pt,
          ]
        \addplot [sharp plot, mark=*, color=blue] coordinates {
          (0, 8.13) (1, 9.68) (2, 9.29) (3, 7.98) (4, 6.48)
          };
        \addplot [sharp plot, mark=*, color=red] coordinates {
          (0, 10.06) (1, 11.29) (2, 10.83) (3, 10.01) (4, 9.74)
          };
        \end{axis}
      \end{tikzpicture}
      \caption{BLEU scores (\%) of different sentence lengths in the test set.}
      \label{ablation_bleu}
    \end{minipage}
  \end{figure}

  We visual the learning curve for comparing the convergence of our method on WMT'16 En-De datasets of different sizes in Figure \ref{learning_curve}.
  One obvious difference in the learning curve is that the performance changes of high-resource datasets (300K/4.5M) during training are more similar to square root function, and low-resource datasets (50K/100K) are more similar to the linear function.
  This phenomenon is consistent with the above experimental results and shows the necessity of calculating the model competence self-adaptively. 
  
  We also observe that our proposed method converges faster and better than the baseline model significantly in all datasets, especially in early phases.
  On the extremely scarce dataset (50K), DCL improves the performance significantly by learning the bilingual data effectively.

  On the other hand, our method only slightly works for high-resource tasks of 4.5M parallel sentence pairs.
  Although DCL improves the convergence speed in early training, the baseline model catches up with enough training steps.
  A possible reason is that a large-scale parallel corpus includes sufficient knowledge and reduces the benefits of the method.
  The weak model trained with small-scale corpus is easy to underfit or overfit, and DCL method improves the performance by highlighting easy samples. 
  However, the strong model trained in high-resource scenarios should pay attention to learning difficult samples.
  We will explore it in the future work.
  
  \subsection{Average Loss}

  As described in section \ref{DCL}, the loss of a training sample indicates whether the NMT model can predict it well. 
  Figure \ref{avg_loss} shows the average loss of all training samples when they are trained different counts on the En-De 50K dataset.
  Our method selects the samples with the fastest loss decline for learning at each training phase, which achieves the lower loss than baseline when training the same counts. 
  It demonstrates learning with more training data is not always beneficial and the better strategy is to select dynamically according to the current model state.

  \subsection{BLEU with Different Sentence Lengths}

  Curriculum learning over-samples the easy samples and one underlying drawback is that it may reduce performance on hard samples due to less training.
  Although sentence length can not represent the difficulty of the training samples accurately, a widely accepted conclusion is that it is more difficult to translate longer sentences.
  We visual the training counts of samples with different sentence lengths  in Figure \ref{training_counts}.
  The training counts of samples in our method is significantly less than baseline, especially for long sentences.
  This also shows that long sentences are also more difficult in our measurement method.

  We divide the test set into different groups according to sentence length and show the BLEU scores of baseline and our method in Figure \ref{ablation_bleu}.
  We observe that our method outperforms the baseline model significantly in translating different length sentences, especially in translating the shorter and longer sentences.
  This demonstrates the organized training process can achieve better translation performance for hard samples even with fewer training times. 

\section{Conclusion}
  
  In this paper, we propose a dynamic curriculum learning (DCL) method to explore the effective use of bilingual data for low-resource NMT.
  We define the difficulty of a training sample by the decline of loss and estimate the model competence self-adaptively based on the performance of the development set.
  Different from previous work, we re-arrange the curriculum once the model is updated, so that the training data with appropriate difficulty is learned by the current model effectively.
  Experimental results show that our method outperforms the strong baselines and several curriculum learning-based counterparts on several low-resource translation tasks.

  DCL only modifies the training strategy without any external data, which has great practical significance for real low-resource scenarios. 
  With a strong baseline model, we can improve the effectiveness of other semi-supervised methods, such as generating the high quality back-translation data. 
  In the future, we will rethinking the existing training startegies and explore the application of DCL methods to more difficult tasks, such as unsupervised learning and model training with pseudo data.

\section*{Acknowledgments}

This work was supported in part by the National Science Foundation of China (Nos. 61876035 and 61732005), the National Key R\&D Program of China (No. 2019QY1801).
The authors would like to thank anonymous reviewers for their valuable comments.

\bibliographystyle{coling}
\bibliography{coling2020}

\end{document}